\documentclass[runningheads]{llncs}
\usepackage{graphicx}
\usepackage{algorithm}  
\usepackage{algorithmic} 
\usepackage{booktabs}
\usepackage{multirow}
\usepackage{setspace}
\usepackage{amssymb}
\usepackage{graphicx}
\usepackage{epstopdf}
\usepackage{subfigure}
\usepackage{multirow}
\usepackage{bm}
\usepackage{amsmath}
\usepackage{cases}
\usepackage{url}
\begin{document}
\title{Highly Efficient Follicular Segmentation in Thyroid Cytopathological Whole Slide Image}
\titlerunning{Highly Efficient Follicular Segmentation}
\author{Siyan Tao\inst{1} \and
Yao Guo\inst{1} \and
Chuang Zhu \inst{1} \thanks{The corresponding author.}\and
 Huang Chen \inst{2} \and
Yue Zhang \inst{3} \and
Jie Yang  \inst{1} \and
 Jun Liu \inst{1}
}
\authorrunning{Tao, Guo, Zhu et al.}
\institute{Beijing University of Post and Telecommunication, Beijing, China
\email{czhu@bupt.edu.cn}
 \and
China-Japanese Friendship Hospital, Beijing, China
 \and Haohandata Technology Co., Beijing, China
}
\maketitle         % typeset the header of the contribution
\begin{abstract}
In this paper, we propose a novel method for highly efficient follicular segmentation of thyroid cytopathological WSIs. Firstly, we propose a hybrid segmentation architecture, which integrates a classifier into Deeplab V3 by adding a branch. A large amount of the WSI segmentation time is saved by skipping the irrelevant areas using the classification branch. Secondly, we merge the low scale fine features into the original atrous spatial pyramid pooling (ASPP) in Deeplab V3 to accurately represent the details in cytopathological images. Thirdly, our hybrid model is trained by a criterion-oriented adaptive loss function, which leads the model converging much faster. Experimental results on a collection of thyroid patches demonstrate that the proposed model reaches 80.9$\%$ on the segmentation accuracy. Besides, 93$\%$ time is reduced for the WSI segmentation by using our proposed method, and the WSI-level accuracy achieves 53.4$\%$.

\end{abstract}
\section{Introduction}
In the past few decades, the incidence of thyroid cancer has increased a lot in many countries \cite{james2018update}. Early and precision diagnosis is the key factor in curing thyroid cancer. Thyroid fine needle aspiration (FNA) achieves exceedingly accurate results in identifying papillary thyroid carcinoma \cite{barbosa2008peripheral}. The clinicians then examine the slides made by the tissue under a microscope and make judgements. However, this judgement is time-consuming and subjective \cite{gopinath2015development}.

It is important to develop fast and objective automatic thyroid cancer diagnosis based on computational tools. In fact, the automatic diagnosis of thyroid cancer usually adopts the Whole Slide Image (WSI), which is generated through an electronic scanner. These WSIs often are in a very large size ($210000\times140000$), which means the direct use of the above schemes to the entire image is impossible due to the great memory usage requirement \cite{samsi2012efficient}. The follicular areas contain the most important information for experts to make diagnosis decision, and follicular segmentation is also a vital step for the automatic diagnostic algorithms. In this paper, we focus on the highly efficient follicular segmentation in thyroid cytopathological WSIs.

\begin{figure}[!htbp]
\centering
\includegraphics[width=0.6in,height=0.6in]{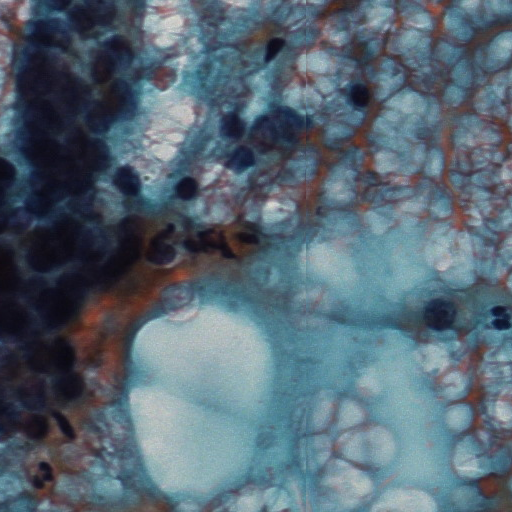}
\includegraphics[width=0.6in,height=0.6in]{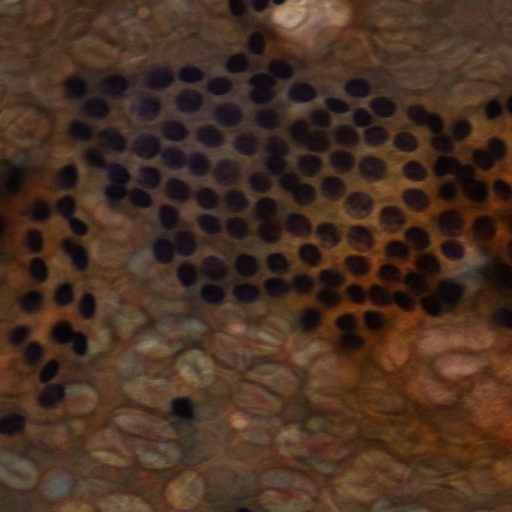} 
\includegraphics[width=0.6in,height=0.6in]{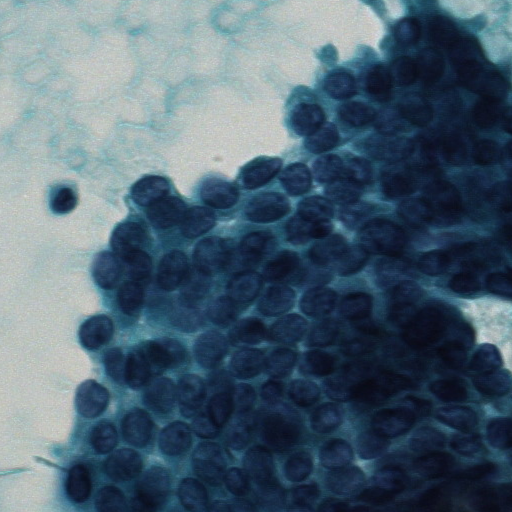}
\includegraphics[width=0.6in,height=0.6in]{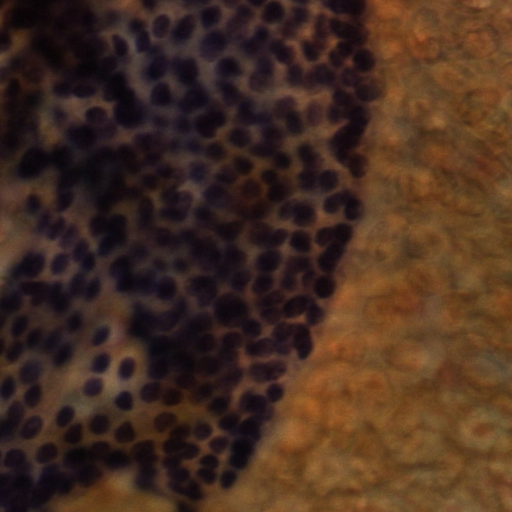}
\caption{ The four images are from different slides.}
\label{wsi}
\end{figure}

Automatic follicular area segmentation for thyroid WSIs faces many challenges due to the following difficulties. Firstly, the data size of the WSI is too large for computers to handle at one time. Secondly, after Pap staining, a large difference between the slides occurs. Figure \ref{wsi} shows the staining of different slices. It can be seen that the stainings of different slides vary greatly. Besides, the follicular cells are usually tightly wrapped by the massive colloid areas, which makes follicular segmentation much harder. 

In this paper, we design a highly efficient accurate follicular segmentation method for thyroid FNA WSIs. We will firstly introduce the hybrid method and the loss function in detail. Secondly, we will experiment with patches and WSIs. Finally, the model will be compared with classic classification models and segmentation models, which will be trained with the same dataset as ours and evaluated with both patches and WSIs.

\section{Related Work}

Traditional machine learning \cite{gopinath2013support,gopinath2015development} methods and deep learning methods \cite{garud2017high,kim2016deep} greatly improve the accuracy of automatic lesion classification in medical areas. Gopinath et al. \cite{gopinath2013support} perform support vector machine(SVM) and achieve a diagnostic accuracy of 96.7\%. B. Gopinath et al. \cite{gopinath2015development} fusion four classifiers and obtain a diagnostic accuracy of 96.66\%. Different from the works mentioned above, Edward Kim et al. \cite{kim2016deep} utilize a deep CNN to the application of thyroid cytopathology classification. Ghosh et al. \cite{garud2017high} present a high accuracy by fine-tuning GoogLeNet \cite{szegedy2015going} in breast FNAC cell sample diagnosis in malignant or benign categories.

Traditional semantic segmentation methods \cite{Preetha2012Image} learn the representation from hand-craft features instead of the semantic features. Recently, CNN-based methods largely improve performance. FCN \cite{Long14fullyconvolutional} is the pioneering work on semantic segmentation by modifying fully connected layers into convolution layers in classification. DeepLab \cite{Chen2015Semantic,Chen2018DeepLab,Chen2017Rethinking} uses dilated convolutions to provide dense labeling and enlarge the receptive field. Semantic segmentation methods have already been used in the pathological image segmentation. Rueckert et al. \cite{Alansary2016Fast} propose a fully automated segmentation framework to identify placental candidate pixels. Cai et al. \cite{Cai2016Pancreas} introduce an image segmentation method based on recurrent neural network.

\section{Method}

\subsection{Dataset Preprocessing}

The dataset used in this paper is the thyroid cytopathological slide provided by a national top-level comprehensive hospital, which is clinical data collected from patients. 

We use the color adjustment method in \cite{Reinhard2002Color} to reduce the influence of the staining. A patch is chosen as a standard of the staining and the other patches are adjusted based on the staining mode of the selected patch.

\subsection{Classification}

In the generated patches of a WSI, only less than 10\% patches contain follicular cells. To label patches and filter the irrelevant patches out, we merge a classifier into the segmentation model.

Patches are divided into three categories: the patches containing the follicular area, the patches of the colloidal area and the patches of the blank non-information area. The patches labeled Follicular are the target patches for the segmentation.

\begin{figure}[!htbp]
\centering
\includegraphics[height= 3.5cm]{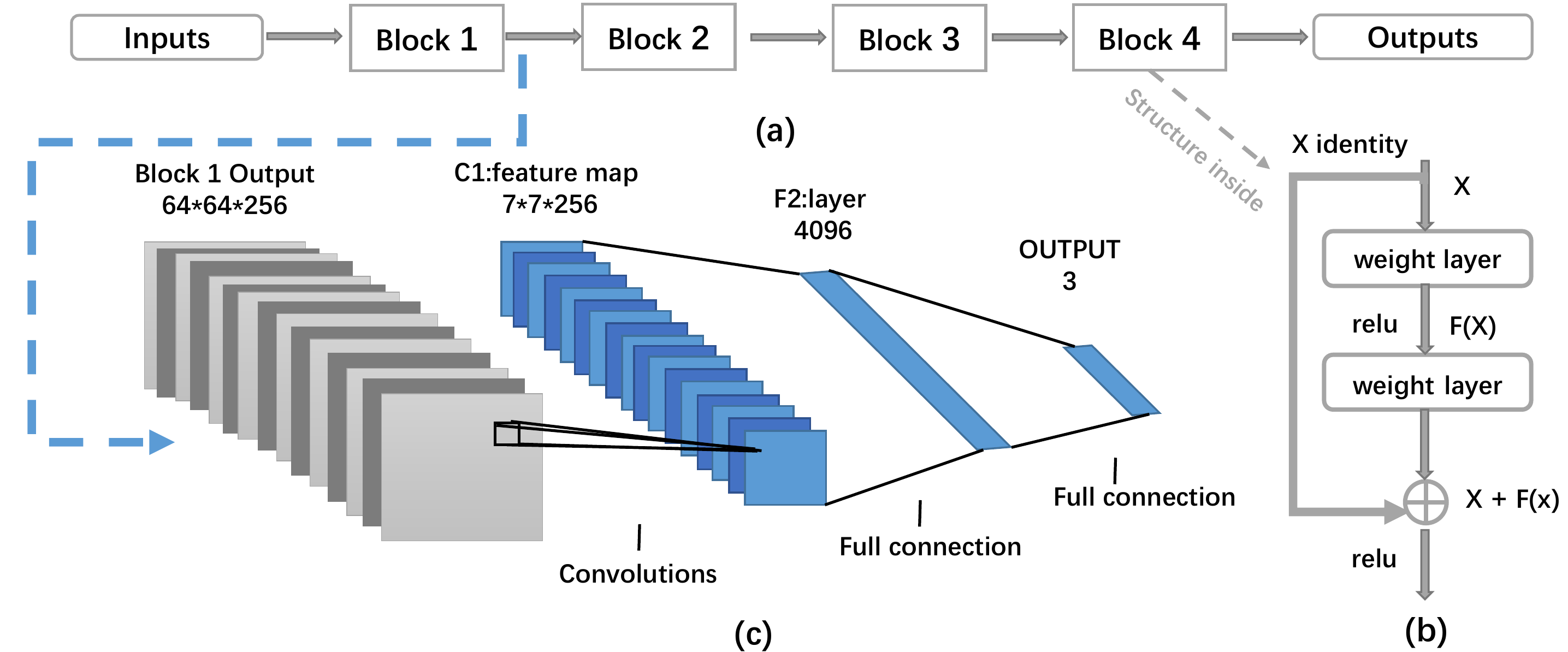}
\caption{(a) The structure of ResNet 101. (b) The basic structure of each block in ResNet 101. (c) The classifier model we propose.}
\label{resnet_class}
\end{figure}

We share the same layers in classifier and segmentation model in order to avoid introducing many parameters. The shared structure is Block 1 of ResNet 101 \cite{He2015Deep}. The structures of ResNet 101 and Blocks are shown in Figure \ref{resnet_class}(a)(b).

We design other layers of the classifier as Figure \ref{resnet_class}(c) shows. The input of the classifier is the output of Block 1 in ResNet 101. A convolution layer and two fully connected layers are added. The final fully connected layer has 3 output nodes which are the same as the category number of the dataset. The loss function of the classification model is the average cross entropy.

\subsection{Segmentation}

\subsubsection{E-ASPP.} The dilated convolutions used in the atrous spatial pyramid pooling (ASPP) are extracted multi-scale information. However, they ignore many relevant detail features which are significant for the thyroid cytopathological WSI dataset. We propose an enhanced ASPP (E-ASPP), which adds precise low scale features to ASPP in order to make up for the deficiencies.

\begin{figure}[!htbp]
\centering
\includegraphics[height= 3.5cm]{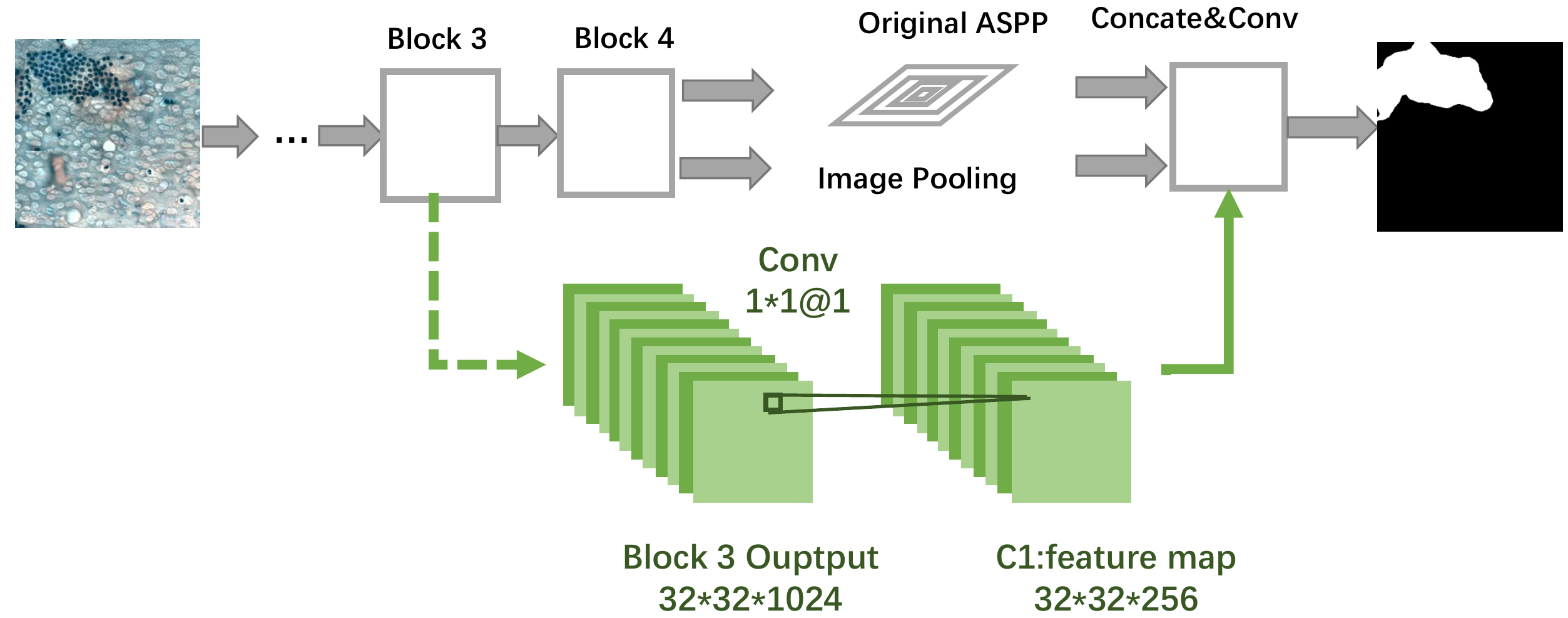}
\caption{The figure shows E-ASPP we propose. The green part adds the low scale features to the original ASPP to offset the deficiencies. }
\label{deep}
\end{figure}

Figure \ref{deep} shows E-ASPP in our method. Beside the structure already existed, we add the low scale features from Block 3 into the original ASPP. E-ASPP offsets the deficiencies of ASPP and improves the accuracy on the follicular segmentation. 

\subsubsection{Criterion-Oriented Adaptive Loss Function.} To lead the model converging much faster, we propose a criterion-oriented adaptive loss function. 

Equation \ref{loss} shows the criterion-oriented adaptive loss function. In a batch, $M$ represents the value of the certain criterion while the denominator is the average cross entropy of patches. It gives a weight to lead the model converging much faster based on the criterion which is used to evaluate the model. $p(x)$ is expected probability distribution that comes from ground truth, $q(x)$ is predicted probability distribution that comes from the prediction of the model. 

\begin{eqnarray}
loss_{seg}= -\frac{\frac{1}{n}\sum_{x}p(x)logq(x)}{M}
\label{loss}
\end{eqnarray}

In this paper, four traditional criteria is used to give $M$ practical meanings: pixel accuracy (pAcc), mean accuracy (mAcc), mean intersection over union (mIoU) and frequency weighted intersection over union (fwIoU) \cite{Long14fullyconvolutional}. They usually are used to evaluate the performance of the semantic segmentation. 

\begin{figure}[!htbp]
\centering
\addtocounter{subfigure}{0}
\subfigure[pAcc]{
\includegraphics[width=1.2in]{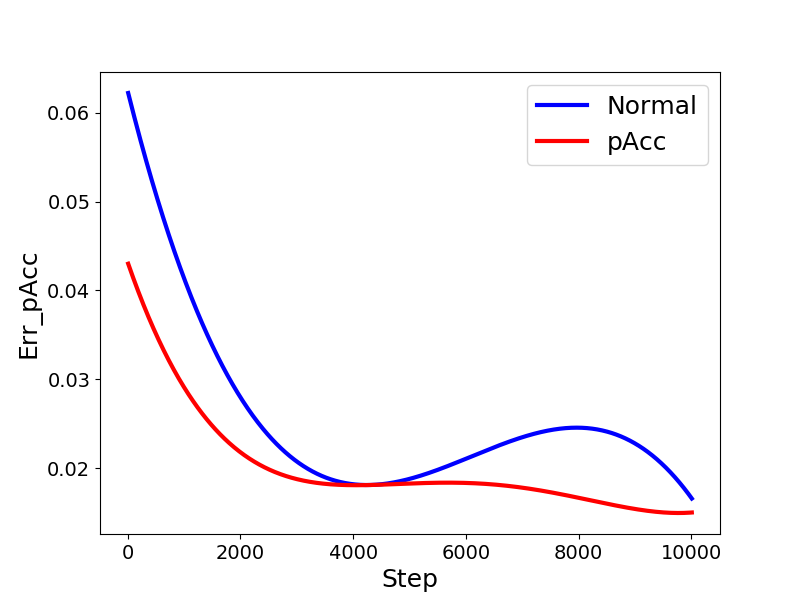}}
\subfigure[mAcc]{
\includegraphics[width=1.2in]{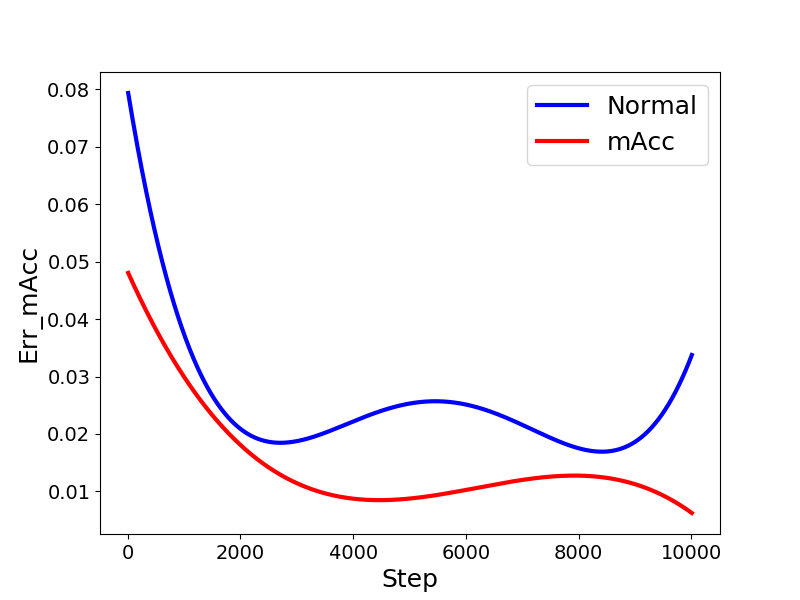}}\\
\subfigure[mIou]{
\includegraphics[width=1.2in]{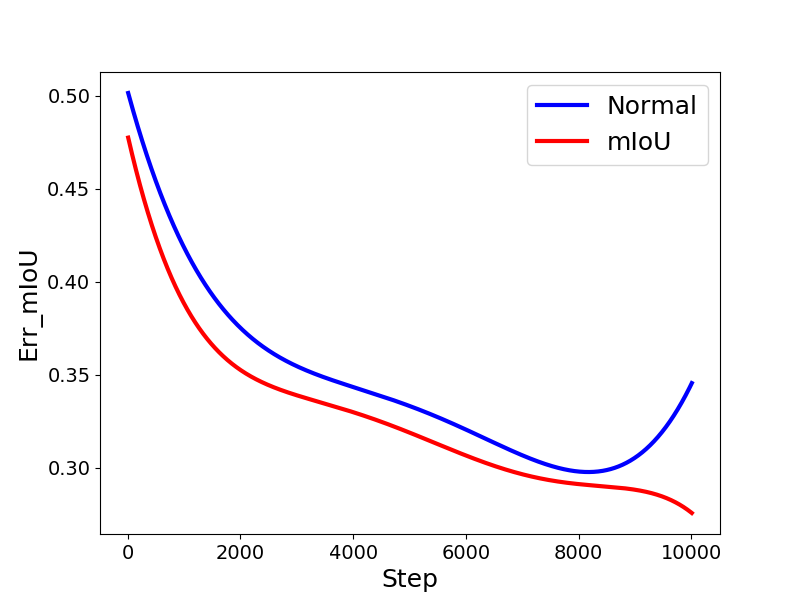}}
\subfigure[fwIoU]{
\includegraphics[width=1.2in]{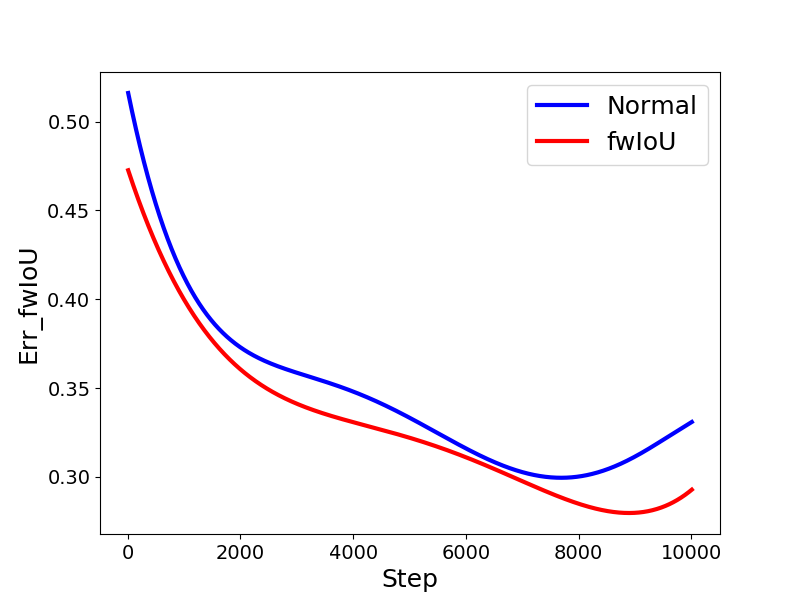}}
\caption{The effects of four different criterion-oriented adaptive loss functions and the cross-entropy loss function.}
\label{lossfunction}
\end{figure}

We compare the effects of criterion-oriented adaptive loss functions for different criteria with the effect of the cross-entropy loss function in Figure \ref{lossfunction}. Under the same number of iterations, the loss function proposed in this paper can make the certain criterion achieve better results faster.

\subsection{Training Method}

We jointly train the hybrid model. As two problems generate two different loss functions, the final loss function is weighted summing of them. The weight can be adjusted according to different situations. In our experiments, the weight is $0.5$.

\section{Experimental Evaluation}

\subsection{Training Environment \& Dataset}
We use Centos 7.0 server to conduct the experiments. The training process uses 2 NVIDIA GTX 1080Ti 12GB GPU (NVIDIA Corporation, Santa Clara, CA) and the NVIDIA Deep Learning GPU Training System (DIGITS 4.0) which has the tensorflow deep learning framework inside. 

The dataset used in this paper contains 15 WSIs. The dataset is divided into two parts: the patch dataset and the WSI dataset. The patch dataset consists of 13 WSIs while the WSI dataset consists of 2 WSIs. We use the patch dataset to train and preliminary test model effect. The WSI dataset is used to test the effectiveness of the hybrid model in practice. 

It is worth noting that all the models in this paper (our model and comparative models) are trained using the thyroid cytopathological image dataset instead of using pre-trained models for fine-tuning.

\subsection{Performance of the Classifier}

To evaluate the classifier objectively, we compare it with classic classification models:LeNet \cite{L1998Gradient}, AlexNet \cite{Krizhevsky2012ImageNet}, GoogLeNet \cite{szegedy2015going}. All the models are trained using thyroid cytopathological image dataset. Table \ref{Acuimg} shows the comparison results through accuracies. Obviously, except GoogLeNet, the classifier we propose has the best performance on this classification problem. The time spent by GoogLeNet is nearly 4 times the time spent by our model while only improving 0.67\% of accuracy. The structure of GoogLeNet is unique and cannot share the same structure with segmentation models. Our classifier finds best balance in accuracy and calculation, which guarantees the efficiency within the scope of fault tolerance.

\begin{table}[!htbp]
\centering
\caption{Accuracy and efficiency of classification models}
\label{my-label}
\begin{tabular}{|c|c|c|c|c|}
\hline
           & LeNet & AlexNEt & GoogleNEt      & Ours           \\ \hline
Follicular & 0.155 & 0.215   & \textbf{0.980} & 0.960          \\ \hline
Colloid    & 0.265 & 0.355   & \textbf{0.985} & 0.980          \\ \hline
Non-info   & 0.535 & 0.600   & 0.995          & \textbf{1.000} \\ \hline
Accuracy   & 0.318 & 0.390   & \textbf{0.987} & 0.980          \\ \hline
Time(s)    & 60.7  & 85.5    & 329.6          & 98.3           \\ \hline
\end{tabular}
\label{Acuimg}
\end{table}

\subsection{Performance of the Segmentation Model}

We experiment with the segmentation structure and compare it with classic segmentation models: FCN, Unet, Deeplab V3. For all the models in this experiment, we train them with the patch dataset to exclude other factors.

\begin{table}[!htbp]
\centering
\caption{Accuarcy of segmentation models on patch and WSI}
\label{my-label}
\begin{tabular}{|c|c|c|c|c|c|c|c|c|}
\hline
\multirow{2}{*}{} & \multicolumn{4}{c|}{Patch}                 & \multicolumn{4}{c|}{WSI}                     \\ \cline{2-9} 
                  & FCN   & Unet  & DeeplabV3 & Ours           & FCN    & Unet   & DeeplabV3 & Ours           \\ \hline
pAcc              & 0.987 & 0.922 & 0.969     & \textbf{0.994} & 0.927  & 0.882  & 0.985     & \textbf{0.987} \\ \hline
mAcc              & 0.867 & 0.513 & 0.743     & \textbf{0.897} & 0.538  & 0.505  & 0.572     & \textbf{0.912} \\ \hline
mIoU              & 0.802 & 0.497 & 0.724     & \textbf{0.809} & 0.512  & 0.495  & 0.503     & \textbf{0.534} \\ \hline
fwIoU             & 0.972 & 0.933 & 0.966     & \textbf{0.979} & 0.972  & 0.875  & 0.984     & \textbf{0.986} \\ \hline
times(s)          & -     & -     & -         & -              & 5350.2 & 4871.5 & 5878.9    & \textbf{756.3} \\ \hline
\end{tabular}
\label{imgAcu}
\end{table}

To evaluate models accurately, we calculate four criteria to compare the performance specifically. The pAcc and the mAcc evaluate models in the pixel level so that we set $M$ as the definition of pAcc in this session. The mIoU and the fwIoU evaluate models in the IoU level so that we set $M$ as the definition of mIoU in this session. Table \ref{imgAcu} shows criteria values of different models. All the criteria perform best with our method. It proves that E-ASPP and the criterion-oriented adaptive loss function are effective.

\subsection{WSI Segmentation of the Hybrid Model}

We experiment with our method and other models with the WSI dataset. To compare model efficiency more fairly, we add data preprocessing to FCN, Unet, and Deeplab V3 \cite{Komura2018Machine}. The preprocessing method is to use the gradient clustering method to filter non-information patches. Table \ref{imgAcu} shows the accuracies and times of the model after adding data preprocessing. 

All the values of accuracy criteria decrease in the WSI dataset since the WSI dataset is more complex than the patch dataset. However, our method performs well in the complex situation, which exists in real medical diagnosis. The time taken for our model to calculate a WSI is much less than the time taken to other models after preprocessing while ensuring accuracy.

\section{Conclusion}
Focusing on the practical problems of thyroid cytopathological diagnosis, we propose a highly efficient hybrid method for the follicular segmentation problem. The hybrid method integrates a classifier into the segmentation model. At the same time, we propose E-ASPP and a criterion-oriented adaptive loss function which have achieved good results in the accuracy in the follicular segmentation. We experiment with the patch dataset and the WSI dataset. The hybrid method significantly improves previous solutions of the follicular segmentation in thyroid cytopathological WSIs and achieves good performance of efficiency and accuracy.

\section{Acknowledge}
This work is supported in part by the Beijing Natural Science Foundation (4182044), Basic scientific research project of Beijing University of Posts and Telecommunications (2018RC11). This work is conducted on the platform of Center for Data Science of Beijing University of Posts and Telecommunications.

\bibliographystyle{splncs04}
\bibliography{ref}
%Author, F.: Article title. Journal \textbf{2}(5), 99--110 (2016)
%Author, F., Author, S.: Title of a proceedings paper. In: Editor,
%F., Editor, S. (eds.) CONFERENCE 2016, LNCS, vol. 9999, pp. 1--13.
%Springer, Heidelberg (2016). \doi{10.10007/1234567890}
%Author, F., Author, S., Author, T.: Book title. 2nd edn. Publisher,
%Author, A.-B.: Contribution title. In: 9th International Proceedings
%on Proceedings, pp. 1--2. Publisher, Location (2010)
%LNCS Homepage, \url{http://www.springer.com/lncs}. Last accessed 4
%\end{thebibliography}
\end{document}